# Enhancing user stance detection on social media using language models: A theoretically-informed research agenda

Prasanta Bhattacharya[1], Hong Zhang[2], Yiming Cao[3], Wei Gao[2], Brandon Siyuan Loh[1], Joseph J.P. Simons[1], Liang Ze Wong[1]

[1] Institute of High Performance Computing (IHPC), Agency for Science, Technology and Research (A*STAR), 1 Fusionopolis Way, #16-16 Connexis, Singapore 138632, Republic of Singapore

{prasanta_bhattacharya, brandon_loh, simonsj, wong_liang_ze}@ihpc.astar.edu.sg

[2] School of Computing and Information Systems, Singapore Management University, 80 Stamford Rd, Singapore 178902

hong.zhang.2022@phdcs.smu.edu.sg, weigao@smu.edu.sg

[3] Department of Computing, The Hong Kong Polytechnic University, Hong Kong SAR, China

yiming.cao@connect.polyu.hk

**ABSTRACT**

Stance detection has emerged as a popular task in natural language processing research, enabled largely by the abundance of target-specific social media data. While there has been considerable research on the development of stance detection models, datasets, and application, we highlight important gaps pertaining to (i) a lack of theoretical conceptualization of stance, and (ii) the treatment of stance at an individual- or user-level, as opposed to message-level. In this paper, we first review the interdisciplinary origins of stance as an individual-level construct to highlight relevant attributes (e.g., psychological features) that might be useful to incorporate in stance detection models. Further, we argue that recent pre-trained and large language models (LLMs) might offer a way to flexibly infer such user-level attributes and/or incorporate them in modelling stance. To better illustrate this, we briefly review and synthesize the emerging corpus of studies on using LLMs for inferring stance, and specifically on incorporating user attributes in such tasks. We conclude by proposing a four-point agenda for pursuing stance detection research that is theoretically informed, inclusive, and practically impactful.

# 1. INTRODUCTION

Social media platforms have become a primary venue for users to connect and express their opinions on emerging issues of interest. Analysing these expression on social media offers substantial benefits across a wide range of applications such as predicting the spread of misinformation and online rumours (Lukasik et al., 2016; Hanselowski et al., 2018; Ma et al., 2018; Umer et al., 2020; Yang et al., 2022, 2024), understanding socio-political phenomena like political debates, elections, and polarization (e.g., Li et al., 2021; Diaz et al., 2022), addressing public health issues (e.g., Glandt et al., 2021), examining environmental outcomes (e.g., Luo et al., 2020), and forecasting stock market returns (e.g., Wang et al., 2020). As a result, the task of inferring the stance conveyed in a social media message or by specific users has gained considerable academic interest in recent years (Aldayel & Magdy, 2019). Drawing from psycho- and socio-linguistics (e.g., "stance triangle"; Du Bois, 2007), stance can be broadly defined as the external expression of one's viewpoint or position on a specific target or topic. It is essential to distinguish stance from related concepts like sentiment or emotions. While both carry information about the author's internal states they differ in their manifestations (Aldayel & Magdy, 2021). For instance, stance is always directed towards a specific target, whereas sentiment might or might not be target-specific. Similarly, sentiment can often be *directly* extracted from surface-level textual or other non-verbal content, while stance may require *indirect* inference through non-target semantic features (e.g., a positive stance towards a political figure *x* might be inferred indirectly from a negative stance towards a rival political figure *y*), or through non-semantic features such as interaction patterns.

Stance detection has been a popular area of research, with recent reviews and studies highlighting its importance (Küçük & Can, 2020; Aldayel & Magdy, 2021; Hardalov et al., 2022a; Alturayeif et al., 2023). Stance detection tasks have also been the subject of important workshop challenges, notably the SemEval 2016 Task 6 (Mohammad et al., 2016). Most research on this topic has focused on social media datasets, mainly derived from X (formerly Twitter), typically framed as target-specific and message-level stance detection tasks. More recent studies have investigated related tasks such as cross-target, target-agnostic, or zero-shot stance detection (e.g., Liang et al., 2021; Wei & Mao, 2019; Zhang et al., 2020; Choi & Ko, 2023; Cruickshank & Ng, 2024). Additionally, there has been a notable expansion in modelling architectures, leveraging pre-trained language models (e.g., BERT-based models; Glandt et al., 2021) and large language models (e.g., Zhang et al., 2024a; Cruickshank & Ng, 2024).

In this position paper, we aim to address two significant omissions in existing work on stance detection and propose a methodical solution to these challenges. The first omission concerns the choice of feature modalities in stance modelling. While prior studies have largely focused on message-level properties (e.g., semantic and interactional features), there is a notable lack of attention to deeper user attributes, including traits, values, and preferences, with only a few exceptions (e.g., Benton & Dredze, 2018; Zhang et al., 2024c). We argue that this omission is particularly unfortunate, since stance, as a construct, is inherently linked to cognitive and socio-political factors, as we elaborate on in the following section. The second omission relates to the choice of task. Most prominent studies on stance detection have focused on *message-level* stance detection tasks, with only a few exploring *user-level* inferences on social media (e.g., Darwish et al., 2020; Samih & Darwish, 2021; Zhang et al., 2024b; Zhang et al., 2023c). However, we argue in this paper that stance is fundamentally a user-level construct. We believe that the popularity of modelling stance at the message level is not due to a deliberate conceptualization of stance as a characteristic of the message, but rather a result of practical constraints in obtaining rich user attributes from social media or self-reported questionnaires, which has limited investigations into user-level stance.

We contend that user-level stance detection is both important and conceptually well-suited to benefit from recent advancements in modelling deeper user attributes (e.g., Zhang et al., 2024c). The popularity of Pre-trained Language Models (PLMs) has opened new opportunities to address aforementioned omissions and constraints. For example, we can now model richer user attributes, such as users' primal beliefs (e.g., Vu et al., 2022) or their moral foundations (e.g., Nguyen et al., 2024), using textual data and use these features to infer their stance on contentious topics. Similarly, the advent of Large Language Models (LLMs) has made it even more accessible to model user personas (e.g., big five personality traits; Jiang et al., 2024) and a wide range of other psychological attributes (e.g., moral foundations, values, etc.) (Bulla et al., 2024; Simons et al., 2024). This can help us infer a more comprehensive profile of social media users, and in turn, their stances on both preexisting and unseen targets. However, to fully leverage the advantages of language modelling in enhancing stance detection tasks, we first need to understand (a) what user attributes, particularly psycho-social attributes, might be relevant in the context of online stance inference, and (b) how recent language models can be utilized to incorporate such user attributes, either through prompting strategies or finetuning techniques.

In the following section, we introduce a formal and theoretically grounded definition of stance by drawing on the topic's multi-disciplinary origins. Specifically, we develop a framework that highlights how stance originates, manifests, and is expressed on public platforms, emphasizing its relevance for designing and extracting stance-related features. In the next section, we review current research on social media-based stance detection, highlighting the various types of tasks (e.g., target-specific vs. cross-target) and the increasing role of language models in this area. We then focus specifically on the potential of using LLMs to infer stances on social media, highlighting the growing evidence of their usefulness. Lastly, we conclude with an agenda for social media-based and theoretically grounded stance detection research, offering insights for computational researchers and platform developers.

## 2. WHAT IS A STANCE?

The expression of stance, as a behaviour, has been extensively studied in the context of stance detection (e.g., Aldayel & Magdy, 2021; Alturayeif et al., 2023) and related tasks (e.g., Ma et al., 2024; Yang et al., 2022; Hardalov et al., 2022a). However, we observe that most prior studies either lack a formal definition of stance or define it very broadly as the position taken by users on a topic of interest. Interestingly, however, the concept of stance has been discussed extensively in other foundational disciplines, such as sociolinguistics, psycholinguistics, communication studies, and philosophy. While definitions emerging from these areas are similar, they emphasize distinct aspects of the definition (e.g., epistemic vs. affective stance, self-focused vs. interpersonal stance). We believe that a systematic analysis of the interdisciplinary origins of stance is crucial, not only for disambiguating the meaning of the term but also as a theoretical guide for developing computational models of stance detection. In this section, we summarize key definitions of stance from various reference disciplines and synthesize these into a simple taxonomy that can inform the selection of data modalities and model design in stance detection tasks.

Our literature search identified 28 distinct usages of the term "stance" across 19 studies spanning a range of disciplines such as sociolinguistics, anthropology, social psychology, philosophy, social computing, and communication studies. Table A1 summarizes these definitions, highlighting the relevant themes and their respective implications for the design of stance detection models. Some of the early definitions come from Biber et al., (1999) and Biber & Finegan (1998) who describe stance as the public expression of attitudes, feelings, or judgments towards an externally defined target (e.g., a message or proposition). Biber et al. (1999), for example, specify that stance involves judgments about the 'certainty, reliability, and limitations' of the target. Stance can also be thought of as a specific philosophical position or perspective of seeing the world (e.g., Teller, 2004), manifested by a 'cluster of attitudes,' as described by Van Fraassen (1995).

These definitions imply that stances need not always be explicitly expressed, even though they often are. In this regard, stance shares conceptual overlaps with other "internal" constructs, such as beliefs and values (e.g., Teller, 2004; Chakravartty, 2004; Boucher, 2014). While some studies have treated stance as a distinct concept from belief (e.g., philosophical stance as an epistemic guide for belief formation; Teller, 2004), others see stance as being coherent with, and inclusive of, beliefs and values (e.g., Boucher, 2014). This suggests that understanding an individual's stance can significantly benefit from an assessment of related psychological constructs, including attitudes, beliefs, cultures, and world views.

There is a highly relevant conceptual connection with attitudes, in particular. One of the oldest and most studied concepts in social psychology (Briñol et al., 2019), attitudes are generally conceived as an enduring evaluation (e.g., positive or negative) of a target (Eagly & Chaiken, 1993). This target could be a person, physical object, or something more abstract such as a moral value. Hence, liking chocolate ice cream is a positive attitude, whereas opposing the death penalty is a negative attitude. While an attitude can be rooted in affective emotions or cognitive beliefs, it is the resulting evaluation which is definitional.

Connecting stances to attitudes can provide greater conceptual clarity on the stance construct. There is a strong similarity between the two, in that they are both valenced evaluations of targets. However, they differ in one important way - stances are generally *public* statements of an evaluation, whereas an attitude is an individual's *private* evaluation. Thus, while a stance may carry information about an attitude (and vice versa), the two things are distinct; the positions people take and the evaluations they privately hold are likely related but not perfectly so. Conceptualizing the constructs in this way helps with better clarifying the existing distinction between *message-level* and *user-level* stances. Specifically, we contend that rather than just a similar phenomenon at different levels, the former are more akin to public evaluations and the latter more like private evaluations (and hence may vary in terms of how these need to be computationally modelled e.g., using distinct feature modalities and modelling architectures). Finally, and perhaps most fruitfully, it suggests that the attitudes literature can be used as a guide for refining the concept of stance, and consequently, for developing improved stance detection models. For instance, prior research has identified a number of consequential aspects of attitudes, such as level of certainty (e.g., Petrocelli et al., 2007), importance to the holder (e.g., Howe & Krosnick, 2017), and level of moral mandate (e.g., Skitka, 2010). While these may not all be directly translatable into stance detection, they can provide a useful guide as to how stances can differ from one another beyond affective distinctions. Consistent with this, recent studies have shown that the incorporation of attributes such as moral foundations can indeed improve stance detection performance (e.g., Zhang et al., 2024c; Nguyen et al., 2024) under specific conditions.

Our review also highlights the interpersonal nature of stance, as defined in Scherer (2005) and Kiesling (2009). While Scherer (2005) defines stance as an 'affective style' (e.g., being polite or warm vs. distant or cold) that one strategically employs in social interactions to achieve the desired outcomes, Kiesling (2009) defines interpersonal stance in terms of the nature of the relationship (e.g., friendly vs. dominating). Beyond the dyadic level, stance has also been conceptualized at a social or normative level, most prominently by Du Bois (2007), who defines it as "an articulated form of social action" employed strategically by social actors using verbal or non-verbal communicative strategies. Thus, we argue that interactional features (e.g., communication styles, interactional network properties, partner characteristics) should also be considered in stance detection tasks, and we see early evidence of this in recent studies that have incorporated social network and interactional features into stance detection models **(**Darwish et al., 2020; Zhang, 2024b**)**.

Stance can also be studied based on the affective vs. epistemic distinction. Affective definitions of stance emphasize feelings or emotions in relation to a target (e.g., Chindamo et al., 2012; Ochs, 1996) and the associated display of this affect (e.g., Goodwin et al., 2012). These definitions align well with how stance is typically used in computational disciplines, focusing on whether an actor or message displays a positive or negative stance towards a certain target. However, as emphasized in this section, stance can also be defined purely in terms of epistemic factors, such as the certainty or reliability with which someone believes a particular assertion (e.g., Kiesling, 2009; Biber et al., 1999), or as a high-level view or perspective of the world that guides belief formation and operationalization (e.g., Teller, 2004). Lastly, and perhaps less relevant to the current discussion, philosophical treatments of stance have also considered ethical and moral dimensions, including the intentional stance, which ascribes intentionality and associated cognitive factors (e.g., attitudes or beliefs) to a target, and the phenomenal stance, which ascribes "felt" states such as emotions and moods to the target (Robbins & Jack, 2006).

We synthesize this discussion on the origins, manifestation, and conveyance of stance using a simple framework as shown in Fig. 1 below. We contend that stance originates as either an epistemic or affective response to a given target (e.g., a person, event, or piece of information). This reaction is then manifested through verbal or non-verbal behaviour (e.g., language use, facial expressions and emotions), interactional styles (e.g., approach vs. avoidance in conversation), psychological changes (e.g., activation of certain beliefs or attitudes), and culture-protective reactions (e.g., positive/negative feelings towards in-group/out-group targets). While these manifestations can remain internal, they are often expressed through behaviour, such as via social media conversations on particular topics. Displays of stance can occur at a personal level (e.g., writing an online post about a political event), an interpersonal level (e.g., commenting positively or negatively on another user's post), or a societal level (e.g., engaging in collective action or social movements).

In the following sections, we highlight how recent advances in computational modelling can be leveraged to model this multimodal manifestation of stance. Specifically, we discuss the role of language models in stance detection tasks and the growing importance of LLMs as a way to augment stance detection models with relevant user-level attributes, as outlined in this section.

Figure 1: A general framework for stance formation, representation, and conveyance.

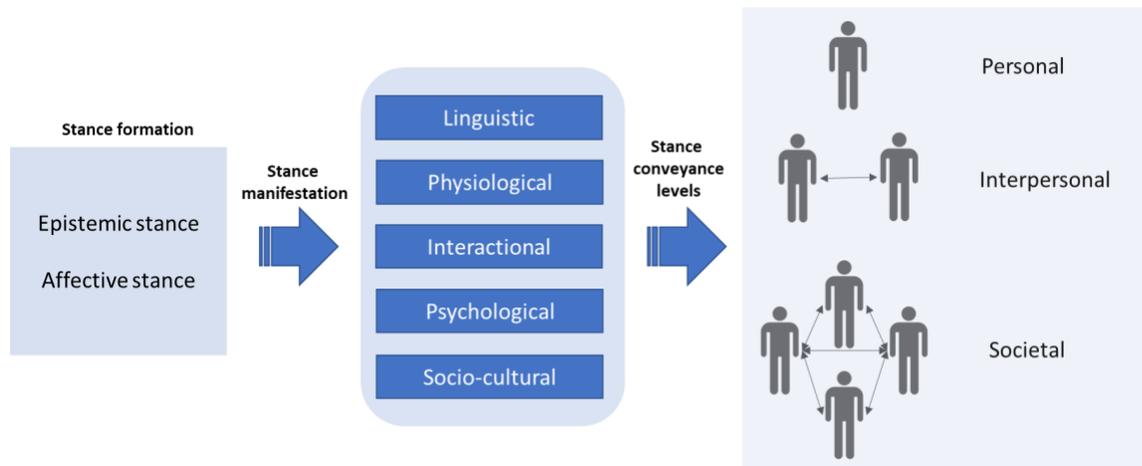

## 3. COMPUTATIONAL MODELS OF ONLINE STANCE INFERENCE

In this section, we summarise the current state of work on developing computational models of online stance detection. As discussed earlier, stance detection is the task of determining the position held by a message or its author towards a specific target, e.g., whether they are in favour, against, or neutral (e.g., SemEval Task 6; Mohammad et al., 2016). While several studies have explored the use of sentiment in stance detection tasks due to their conceptual similarities, Aldayel et al. (2021) contend that there are unique aspects of stance not captured by sentiment. In other words, sentiment alone is often an insufficient indicator of stance. For example, considering Hillary Clinton as a target, both messages, "I am sad that Hillary lost this presidential race" and "I am happy that Trump won this presidential race," exhibit a favourable stance towards Clinton, despite displaying conflicting sentiments.

With the proliferation of social media, platforms such as X (formerly Twitter) have become popular venues for users to express opinions on diverse issues and targets. This has resulted in a wealth of publicly available opinion and stance datasets (Aldayel & Magdy, 2019; Conover et al., 2011; Kwak et al., 2010; Wang et al., 2012; Li et al., 2015). For example, the SemEval Task 6 (Mohammad et al., 2016) competition featured two stance detection tasks based on a Twitter-based dataset. In Task A, participants were tasked with building a supervised stance detection model to detect stance towards specific targets, including "Atheism", "Climate Change is a Real Concern", "Feminist Movement", "Hillary Clinton" and "Legalization of Abortion". In Task B, participants were required to build a weakly supervised stance detection model for the target "Donald Trump". This competition and dataset spurred significant interest in stance detection research, evidenced by a sharp increase in the number of stance related studies since 2016 (Alturayeif et al., 2023).

We believe that the availability of high-quality stance detection datasets offers a unique opportunity for computational social scientists who are interested in studying human behaviours and opinions around key topics or events (Coppersmith et al., 2018; Tyshchuk & Wallace, 2018; Lappeman et al., 2020; Zhang et al., 2023a; Alturayeif et al., 2023). Over the past decade, there has been a surge of studies aimed at extracting stance from language used on social media to better understand users' opinion or attitudes. However, as previously discussed, many of these studies are limited by (a) their focus on message-level stances rather than user-level stances, and (b) a lack of focus on important user attributes (e.g., demographic attributes, psycho-social factors, political preferences, etc.) that might significantly contribute to the expression of online stances.

### 3.1. Target-specificity of stance detection tasks

Stance detection models can be distinguished by their specificity to certain focal targets. In both target-specific and multi-target stance detection, models are trained to identify user or message-level stances towards a predefined set of targets, such as specific topics or individuals (e.g., Aldayel & Magdy, 2021; Hasan & Ng, 2013; Walker et al., 2012). Multi-target stance detection is often considered advantageous because it leverages shared information between various related targets, enhancing the stance detection process compared to other approaches that require target-specific model re-training. However, both methodologies are confined to inferring stances only on the set of targets they have been trained on. Indeed, research by Sobhani et al. (2017) demonstrates that the effectiveness of multi-target stance detection depends on how closely related the targets are. In most real-life applications, the wide range of targets presents a formidable challenge, and acquiring adequate data for training target-specific or multi-target models is frequently infeasible, underscoring a fundamental challenge with these approaches.

In contrast, zero-shot stance detection models are designed to operate on targets that might be *unseen* during model training. This capability is particularly important in real-world scenarios where stance targets might be unknown, emerging, or too many to model efficiently. Zero-shot stance detection models address this problem by allowing stance detection to be performed on targets that were not necessarily included in training, and this has substantially enhanced the scalability of stance detection applications (e.g., Allaway et al., 2021; Liang et al., 2022; Lin et al., 2024). The rise of LLMs, in particular, has significantly boosted the popularity of zero-shot stance detection models, as we discuss next in Sec. 4.

### 3.2. Emergence of online stance detection models

Computational models for online stance detection can be broadly distinguished by their modelling architecture, learning process, and data requirements. In this section, we provide a brief overview of recent progress in each category.

### 3.2.1 Traditional machine learning (ML)

Traditional ML models were prevalent in early stance detection research and continue to serve as effective baseline methods in recent studies. These models are generally supervised and require carefully curated features to classify stance into distinct classes. For example, in SemEval 2016 task A, a Support Vector Machine (SVM) classifier using n-gram features was used as a baseline model (Mohammad et al., 2016). Similarly, Zeng et al. (2016) studied rumour stances using logistic regression, Gaussian Naïve Bayes, and random forest models trained using Twitter mentions, hashtags, LIWC features (Tausczik & Pennebaker, 2010), sentiment, n-grams, and part-of-speech features. In another approach, Dey et al. (2017) proposed a two-phase method using an SVM. In the first phase, an SVM model was trained using weighted MPQA subjectivity-polarity classification and WordNet-based potential adjective recognition features to identify tweets with a neutral stance towards the target. In the next phase, a second SVM model was trained on SentiWordNet and MPQA based sentiment classification, frame semantics, and other features (e.g., presence of target in tweet, n-gram features) to classify non-neutral tweets into "favour" vs. "against" classes.

Notably, user-level stance detection using such traditional models remains relatively understudied, with some recent exceptions such as Samih & Darwish (2021) who used an SVM to classify Twitter users' stance. Common user-level features in stance detection tasks include attributes that are easily extractable from either the content (e.g., tweets) or the public user profile. For instance, stance detection models have utilized the number of followers, user registration time, verification status, social network structure, and retweet behaviour (Samih & Darwish, 2021; Benton & Dredze, 2018; Xuan & Xia, 2019). However, there is a clear lack of research that leverages deeper user attributes, such as users' psychological states, to infer users' stance.

### 3.2.2 Deep learning (DL)

DL models have gained prominence in stance detection tasks, particularly with the advent of recurrent neural networks (RNNs) and, more specifically, long short-term memory (LSTM) networks (e.g., Augenstein et al., 2016). These models have been used across NLP tasks, including stance detection. LSTMs can capture long-distance dependencies and effectively address the gradient vanishing and exploding problems that can affect vanilla RNNs (Hochreiter & Schmidhuber,1997). In their study, Du et al. (2017) introduced a Target-specific Attention Neural Network (TAN) using a bi-directional LSTM model to capture information from text, along with a fully connected network as a target-specific attention selector. This model outperformed several strong baselines on both English and Chinese datasets. Similarly, Benton & Dredze (2018) employed a gated recurrent unit (GRU) for stance detection. In this model, the input text was first encoded with GloVe embeddings pretrained on Twitter data (Pennington et al., 2014), which were then processed by GRU to produce the final hidden state activations. The model then classified stance based on a convex combination of these activations.

The flexibility of deep learning models allows for complex architectures involving multiple models and tasks, allowing studies to focus on multi-model and multi-task frameworks for stance detection. For example, Mohtarami et al. (2018) proposed a novel memory network with five components. The *input representation component* converted each document input into a set of evidence for stance detection, leveraging an LSTM for effective memorization of input and a CNN to capture the local interaction between words. The *inference component* computed semantic similarity between claims and evidence, while the *memory and generalization components* controlled information flow and updates within the network. Finally, the *output representation* component assessed the relative perspective of a document concerning a claim, and the *response and output generation component* inferred the final stance of the document. Additionally, Li & Caragea (2019) proposed a multi-task framework with target-specific attention mechanism for stance detection that used sentiment classification as an auxiliary task. In other work, Allaway et al. (2021) developed a Topic-Adversarial Network for zero-shot stance detection, using adversarial learning to generalize across topics.

Despite the advancements in deep learning, only a limited number of studies have applied these models to *user-level* stance detection tasks. The few that exist have leveraged users' historical tweets and their social network neighbours' tweets to construct user-level features (Jia et al., 2022; Zhu et al., 2020). In contrast, *tweet-level* and DL-based stance detection models have incorporated user-level features such as tweet content, social network information (e.g., count of friends and followers), the volume of posts, public lists the user belongs to, and verification status. While most studies on both tweet-level and user-level stance detection rely on user profiles and tweets to create user-level features, modelling stance with deeper user-level attributes remains an understudied area.

### 3.2.3 Pre-trained Language Models (PLMs)

PLMs have been widely used and have achieved state-of-the-art results in many NLP tasks since the introduction of the Bidirectional Encoder Representations from Transformers (BERT) model (Devlin et al., 2019). Through pre-training on a large corpus, these models learn the general semantics of language and can subsequently be fine-tuned for specific downstream tasks. In a study focused on developing DL models for stance detection, Li et al. (2021) created the P-Stance dataset, which contains 21,574 labelled tweets about three political figures from the 2020 US presidential race. Their study showed that BERTweet, a finetuned BERT-based model, significantly outperformed baseline models in both in-target and cross-topic stance detection tasks. In related work, Liu et al. (2022) employed stance detection as a tool for analysing political ideology. Specifically, they fine-tuned RoBERTa on the BIGNEWS dataset and incorporated triplet loss along with masked language models, resulting in their POLITICS model outperforming a baseline RoBERTa model across three datasets. Similarly, Zhang et al. (2023a) used a BERT-based distantly supervised model to detect the stance of 183,848 Twitter users towards three related targets in the Connected Behavior dataset, testing the model's capability to predict users' electoral decisions based on past behaviours. Given the strong performance of PLMs in language classification and understanding tasks, Samih & Darwish (2021) and Gambini et al. (2022) used users' timeline tweets for user-level stance detection using BERT, BART, and XLM-RoBERTa based models. Similarly, Zhang et al (2024c). used BERTweet to combine tweet and moral foundation features to enhance stance detection at both tweet and user levels. To our knowledge, this study is among the first to incorporate deeper user attributes in a user-level stance detection task.

Labelling stance datasets is a labour-intensive task that requires annotation of large training datasets. Moreover, accurate labelling often requires domain-specific knowledge about the target, making the annotation process even more challenging. Unsupervised learning (UL) models present a viable solution to this challenge, allowing researchers to classify text or users into stance categories without prior labels. For example, Darwish et al. (2020) incorporated techniques like Fruchterman-Reingold force-directed (FD) graph drawing algorithm (Fruchterman & Reingold, 1991), t-distributed Stochastic Neighbour Embedding (t-SNE, Van der Maaten & Hinton, 2008), and Uniform Manifold Approximation and Projection (UMAP) for dimensionality reduction (McInnes et al., 2018), followed by clustering methods such as Density-based Spatial Clustering of Applications with Noise (DBSCAN; Ester et al., 1996) and Mean Shift to develop an unsupervised framework for vocal Twitter users. Their feature set included message-level characteristics such as the cosine similarity for tweets, hashtags, and retweet accounts. Their proposed framework proved effective for active users, achieving 98% purity for the most active 500 or 1,000 users in a dataset of 250k tweets. In related work, Rashed et al. (2021) proposed an unsupervised stance detection framework for Turkish data that utilized Google's CNN multilingual universal sentence encoder and projected the data into a lower dimensional space for clustering.

Broadly, UL models have used discriminating patterns of social media activity, based on tweets, retweets, and hashtags, to effectively classify users into distinct groups reflecting different stances (Samih & Darwish, 2021; Darwish et al., 2020; Zhou & Elejalde, 2024; Rashed et al., 2021). Past studies, such as Aldayel et al. (2019), have also used user-level network features, such as interactions, preferences, and connections, in *tweet-level* stance detection tasks. However, and as with traditional and prior deep learning models, the use of deeper user-level attributes within UL models remains untested.

## 4. ENHANCING STANCE DETECTION USING LARGE LANGUAGE MODELS

LLMs are rapidly emerging as viable alternatives to traditional machine learning models and earlier pretrained language models (Zhao et al., 2023). We argue that LLMs are particularly well-suited to address the gaps in stance detection research for two main reasons. First, there is growing evidence that LLMs can act as effective *feature generators* for a range of textual features. For instance, they can infer deep user-level attributes such as user values or beliefs from social media-based text, which can then be incorporated as additional feature sets in existing stance detection models. Second, LLMs can function as *stance detectors* by utilizing user-level and message-level attributes flexibly (e.g., via prompting strategies) to infer stance on a given target in a zero- or few-shot manner (e.g., Cruickshank & Ng, 2024). While prior PLMs offered some capabilities in this regard, we argue that implementing and customizing these functionalities is considerably easier with currently available LLMs. In this section, we discuss recent evidence on the potential of LLMs to enhance stance detection, and particularly user-level stance detection tasks.

### 4.1. Can LLMs effectively detect stance?

With the recent development and popularity of LLMs in related tasks, researchers have begun to explore their utility for stance detection. We reviewed a total of 23 recent studies (see Table A2) which employed LLMs for stance detection tasks, aiming to understand (a) the key approaches and techniques (e.g., novel prompting schemes, multi-agent modelling, fine-tuning, etc.) used to model stance, and (b) the potential for incorporating deep user attributes as discussed in previous sections. Our findings reveal that LLMs have been used in stance detection research through a range of methods and strategies, as summarized below.

**LLM-enhanced modelling pipelines.** LLMs have often been integrated into stance detection pipelines to generate knowledge-enhanced feature sets for downstream models and tasks (e.g., Lin et al., 2024; Ding et al., 2024a; Zhang et al., 2024a; Dong et al., 2024; Gatto et al., 2023; Wagner et al., 2024; Li et al., 2024; Sharma et al., 2023; Zhang et al., 2024d). For example, Zhang et al. (2024a) included LLM-driven keywords, implied emotions, rhetorical devices, and stance reasoning into a BART model for stance detection. Similarly, Ding et al. (2024a) proposed an attention-based prompt-tuning stance predictor that elicited and fused diverse perspectives from LLMs. In other work, Gatto et al. (2023) integrated Chain of Thought (CoT) reasoning generated by LLMs into a RoBERTa-based stance detection pipeline. LLMs have also been effective at extracting knowledge from Twitter hashtags. For example, Dong et al. (2024) employed a zero-shot prompting method to extract hashtag knowledge, generated hashtag embeddings with BERT, and integrated them into a fusion network for stance detection. Similarly, Li et al. (2024) developed a calibration network that incorporated sample sentences, LLM stance judgments, and rationales to produce debiased stance outputs. Wagner et al. (2024) leveraged LLMs to generate synthetic data for augmenting a fine-tuning dataset and identified the most informative samples for manual labelling to train a BERT-based stance detection model. Lastly, Sharma et al. (2023) demonstrated that combining an ensemble of LLMs through a weighted sum of their predictions outperformed few-shot predictions using the LLaMA-2 model (Touvron et al., 2023).

**Novel prompting schemes.** While many studies have focused on enhancing model pipelines using LLMs, others have explored alternative methods leveraging effective prompting schemes to infer stances. The CoT prompting scheme has emerged a popular technique for stance detection (e.g., Gatto et al., 2023; Cruickshank & Ng, 2024) and has demonstrated superior performance (Cruickshank & Ng, 2024). However, concerns regarding the logical coherence of generated knowledge persist (Zhang et al., 2023c; Ding et al., 2024b), leading a few recent studies to address this challenge. For example, Zhang et al. (2023c) introduced the Logically Consistent Chain-of-Thought (LC-CoT) prompting scheme, which integrated if-then logical structures to enhance the relevance and coherence of LLM outputs in stance detection tasks. Similarly, Šuppa et al. (2024) enhanced few-shot prompting by selecting training examples that closely resembled input samples based on an embedding-based similarity computation.

**Fine-tuned LLMs.** A few recent studies have attempted to fine-tune LLMs, using either popular stance detection datasets or using data from related tasks like argument generation, to enhance their performance (Gül et al., 2024; Yang et al., 2024; Kang et al., 2023; Reuver et al., 2024). For example, Gül et al. (2024) illustrated that fine-tuning LLMs for specific stance detection tasks significantly improved their overall effectiveness, assessed using the $F_{avg}$ score, making them more effective than zero-shot or few-shot approaches. Yang et al. (2024) employed a reinforcement learning-based label selector to choose high-quality examples for fine-tuning stance detection LLMs. Fine-tuning has also been used to incorporate rich user attributes. For example, Kang et al. (2023) proposed a value-injected LLM which integrated targeted value distributions through argument generation and question answering. These value-injected LLMs were then used to predict opinions and behaviours of individuals who share similar distributions of core human values.

**LLMs as multi-agent systems.** Although LLMs are trained on large and diverse datasets, they can occasionally struggle to apply target-specific knowledge effectively (e.g., Lan et al., 2024). To address this, researchers have explored using LLMs as multi-agent systems, where different LLMs act as specialized agents, each offering distinct analytical perspectives (Lan et al., 2024; Wang et al., 2024). For instance, Lan et al. (2024) assigned roles such as linguistic expert, domain specialist, and social media analyst to generate multifaceted explanations from text. A separate LLM then generated a link between these texts and the implied stance, with a final LLM consolidating everything to determine the overall stance. Further, Wang et al. (2024) critiqued previous methods for relying on *static* experts and proposed a dynamic selection process that prompts LLMs to choose from a pool of manually curated experts to reason about stance.

**LLM evaluations using crowd-sourced annotations and other models.** Researchers have also evaluated the performance of LLMs on stance detection tasks to explore their potential to replace crowd-sourced annotations or serve as benchmarks for new models (Aiyappa et al., 2024; Ziems et al., 2024; Cruickshank & Ng, 2024; Li et al., 2024; Gatto et al., 2024; Mets et al., 2024). For instance, Li et al. (2024) compared LLMs with human annotations in stance detection tasks, revealing close alignment in performance. Notably, they highlight that the instances where LLMs falter often mirror challenges faced by human annotators as well. In their evaluative study, Cruickshank & Ng (2024) compared various LLMs and prompting strategies, showing that LLMs outperformed supervised encoder-decoder models when using specific prompting schemes like few-shot and zero-shot CoT. However, Niu et al. (2024) demonstrated recently that a conversational stance detection model built on BERT with a global-local attention layer outperformed LLMs using in-context learning based on a single demonstration sample.

The literature on LLMs in stance detection is still nascent but quickly evolving. Our review of current studies, as summarized in Table A2, shows increasing evidence that LLMs can significantly enhance stance detection tasks. However, challenges remain, including a lack of context-specific knowledge (e.g., domain expertise, cultural references, and social media linguistic styles) and difficulties in understanding implicit reasoning. Unsurprisingly, some studies employing LLMs for stance detection have also reported occasional underperformance compared to non-LLM baselines (e.g., Zhang et al., 2022; Zhang et al., 2023b). To address these challenges, researchers have proposed a number of strategies, such as utilizing LLMs as multi-agent systems that can fuse comprehensive analyses from diverse perspectives. However, beyond performance considerations, recent studies (Pit et al., 2024; Rozado, 2024) have also investigated inherent biases (e.g., political or occupational biases) in LLMs. For example, an LLM with a political slant might exhibit a tendency to align responses with liberal or conservative ideologies. Importantly, Rozado (2024) demonstrated that supervised fine-tuning with even small amounts of politically aligned data can sway an LLM's orientation.

Similarly, Pit et al. (2024) illustrated how LLMs perceive different occupational roles politically. They found that the Llama-2 and GPT-4 models consistently classified roles like Parent and School Administrator as left-leaning, whereas roles such as University Professor and Textbook Publisher were deemed politically neutral. These findings underscore the importance of retaining neutrality in prompt language when utilizing LLMs for stance detection.

**4.2. Encoding user attributes in stance detection LLMs**

While recent studies have explored the effectiveness of using LLMs in stance detection, as discussed in the previous section, the integration of stance-relevant user attributes with LLMs remains understudied. However, early evidence suggests that this could be a promising area for exploration (e.g., Zhang et al., 2024c; Nguyen et al., 2024; Rezapour et al., 2021). There are two broad approaches for incorporating user-level information into LLM-based stance detection models. The first involves injecting user attributes (e.g., human values) directly into LLMs, while the second prompts LLMs to consider user-related information while inferring stance.

**Augmenting LLMs with human values and moral foundations.** Psychological studies on human values and morality indicate that Individuals possess core values and moral foundations that guide their daily decisions and behaviour (Stern et al.,1999; Graham et al., 2013). For instance, Schwartz's value theory (Schwartz, 2012) identifies ten such values—such as security and achievement—that are pivotal to human life. Similarly, the Moral Foundations Theory (MFT) posits that our moral and ethical judgments are based on psychological foundations, which guide not just individual behaviour but also institutional and societal norms across cultures (Graham et al., 2013). The original MFT proposed five foundations: care/harm, fairness/cheating, authority/subversion, sanctity/degradation, and loyalty/betrayal, with additional dimensions like equality, proportionality, liberty, honour, and ownership added later.

We believe that LLMs that are capable of inferring such value systems from available user-level information are likely to make better inferences on downstream tasks, such as user-level stance detection. Moreover, if researchers already know psychological attributes, such as moral foundations and values, they can be incorporated into the LLM via prompting or fine-tuning to improve its performance on downstream tasks. For example, Kang et al. (2023) demonstrated that integrating human values with an LLM via fine-tuning enabled it to predict the opinions and behaviours of individuals who share similar distributions of core human values. Similarly, Zhang et al. (2024c) found that incorporating moral foundation features into the LLM substantially improved modelling accuracy on a range of stance detection tasks, both at the message and user levels, across multiple datasets.

**Augmenting LLMs with sentiment and emotions.** Stance has often been studied in the context of affective stance, making it unsurprising that the emotional tone of an individual serves as an important user attribute for augmenting stance detection performance. Recent research has discussed competing perspectives on how to incorporate emotional tone into LLM-based approaches. The first perspective advocates for prompting LLMs to analyse or consider sentiment or emotional tone in text while inferring user-level stances, particularly in social media-related contexts (Lan et al., 2024; Zhang et al., 2024a). For example, Lan et al. (2024) prompted LLMs to take on the role of a "social media veteran" to analyse the emotional tone in text, revealing underlying sentiments and the authors' feelings. Similarly, Zhang et al. (2024a) prompted the LLM to analyse implied emotions and rhetorical devices, which can serve as useful indicators of the expressed stance. The second and competing perspective cautions against the biases introduced by excessive reliance on sentiment in LLM-based stance detection (Li et al., 2024). In their study, Li et al. highlighted that sentiment information can distort stance judgments and proposed the use of counterfactual augmented data to mitigate these challenges. Specifically, they prompted LLMs to rephrase sentences and targets with varied sentiments while preserving stance, which helped train a calibration network for debiased stance outputs.

While studies on utilizing deep user attributes, such as psychological states of users, in LLMs for stance detection are limited, the few that exist show promise in this direction. Given their extensive pre-existing knowledge base and flexibility in incorporating additional information via prompting or finetuning, LLMs offer a unique opportunity for stance detection researchers to incorporate a rich set of user-level attributes, as discussed in Sec. 2. These attributes can either be inferred directly using the LLM (e.g., inferring moral foundations; Nguyen et al., 2024) or generated through alternative approaches (e.g., using a lexicon or a pre-trained model) and then incorporated into a stance detection LLM (e.g., moral foundations using eMFD; Zhang et al., 2024c). In the following section, we propose an integrative agenda for stance detection research that combines methods and insights from existing research with new possibilities unlocked by using LLMs.

## 5. AN INTEGRATIVE AGENDA FOR FUTURE RESEARCH ON USER-LEVEL STANCE DETECTION

Stance detection has emerged as an important language modelling task, attracting considerable recent interest through the development of state-of-the-art models, new datasets, and systematic reviews on the topic. However, as we highlight in this paper, user-level stances are rarely defined or conceptualized adequately in empirical studies. From a computational perspective, a better understanding of the theoretical underpinnings of stance can lead to improved stance detection models and user-level stance datasets. We also emphasize that the emergence of LLMs offers fresh opportunities to address these limitations and challenges. In this section, we present a set of questions, directions, and opportunities that researchers can consider in future research.

### 5.1. Understanding the psychological foundations of stance formation and conveyance

To develop better user representations for stance detection tasks, it is essential to have a clearer understanding of the user attributes that contribute to stance formation and conveyance. Psychological attributes, in particular, represent a fertile yet understudied area in this field. We have highlighted in Appendix A1 certain themes and constructs that are closely related to stance formation (e.g., beliefs) and stance conveyance (e.g., emotions). However, there is a need to systematically investigate the link between relevant psychological attributes (e.g., salient attitudes or beliefs) and users' stances on specific targets.

Related studies have highlighted this as one of the key challenges in stance detection tasks. For example, the domain and cultural specificity of social media discussions can complicate stance detection. A message might be positively stanced in one domain and/or for users from a specific group, while appearing negatively stanced in other cases (Alkhalifa & Zubiaga, 2022). Clearly, access to richer information about users' socio-cultural backgrounds and the context of conversation could help mitigate some of these challenges. Other studies have similarly called for incorporating user-level information, where available, on diverse demographic attributes (Luo et al., 2020) and traits (Magdy et al., 2016) to improve accuracy in stance detection tasks.

We believe that the users' psychological attributes are valuable for constructing rich user representations, but accessing this data can be challenging due to access limitations and reporting biases. As a result, there are relatively few studies that incorporate such user-level attributes. However, we believe that the growing popularity of LLMs in acquiring or inferring psychological attributes, such as personalities (e.g., Jiang et al., 2024) and moral foundations (Nguyen et al., 2024) is likely to benefit related research on downstream tasks, such as user-level stance detection. Recent studies have demonstrated the utility of incorporating such psychological information to enhance stance detection performance (Zhang et al., 2024b).

**5.2. Need for improved data resources**

Since the popular SemEval Task 6 competition, the availability of publicly accessible and stance-labelled social media datasets has grown considerably (Alkhalifa & Zubiaga, 2022; Alturayeif et al., 2023). However, as mentioned earlier, there is a notable lack of datasets that incorporate a rich set of user attributes for enhancing performance in user-level stance detection tasks. Past studies have discussed the potential of user-level features, notably through user embedding models, to improve stance detection at the message level (e.g., Benton & Dredze, 2018). In these studies, user representations are generally derived from the content they generate, which complicates the task of decoupling the effects of user attributes from the content attributes.

Researchers have also highlighted the usefulness of incorporating contextual information from the users' profile or their activity (e.g., retweets and conversation thread structure) in stance detection tasks, as well as associated heterogeneities in model performance (e.g., Hardalov et al., 2022a; Aldayel & Magdy, 2019; Darwish et al., 2020; Zubiaga et al., 2016). Curating not just larger but also 'thicker' datasets that include useful user information can greatly improve stance detection performance for both message- as well as user-level tasks. It is worth noting that many of these attributes can also be inferred using LLMs (e.g., Nguyen et al., 2024; Simons et al., 2024) and subsequently validated through user studies.

Another crucial consideration is the diversity and representativeness of stance datasets. Recent studies have called for larger benchmark datasets, covering a broader set of topics (beyond social and political issues) and including more diverse languages (Hardalov et al., 2022b; Aldayel & Magdy, 2021). The study by Alturayeif et al. (2023) found that only 15% of the reviewed studies or datasets were in languages other than English. Some recent attempts have aimed to address this challenge via large-scale and multi-lingual evaluations of stance detection (e.g., Darwish et al., 2020; Vamvas & Sennrich, 2020; Hardalov et al., 2022b). We believe that this is an important area for future work as it can lead to the development of more inclusive datasets and task/data challenges. The recent Stance Detection in Arabic Language Shared Task, organized as part of ArabicNLP 2024, is a promising step in this direction (Alturayeif et al., 2024).

Lastly, there is also a pressing need for longitudinal assessments of stance across contexts to better understand how user stances evolve over time (Alturayeif et al., 2023), and in response to specific events (e.g., reactions to terror attacks in Paris; Magdy et al., 2016). While a few stance detection datasets cover relatively long time periods, there is a lack of datasets that capture changes in user-level stance at multiple and periodic intervals for specific targets. Since social media data allows for longitudinal assessments of user behaviour both in terms of content generated and ties formed, stance detection in the aftermath of critical events can provide valuable insights for governance and policy (Aldayel & Magdy, 2021).

### 5.3. Emerging stance detection tasks

While most studies have focused on stance detection for pre-specified targets, the related challenge of stance *prediction* for new or unseen targets is equally important (Aldayel & Magdy, 2021). This is especially true when (a) the new targets are unrelated to those on which the stance detection model has been pre-trained, and/or (b) there is limited access to target-specific content for model fine-tuning. Stance prediction can be thought of as an extension of multi-target and cross-target stance detections tasks, where the datasets comprise messages labelled for multiple targets (e.g., Sobhani et al., 2017). Such tasks often make use of transfer learning techniques, allowing a model trained on a specified set of targets to make inferences about other targets. We believe that both cross-target as well as target-agnostic stance detection can be effectively enabled using LLMs, with early evidence from zero- and few-shot prompting techniques showing potential (e.g., Loh et al., 2024).

Moreover, since stances often reflect the psychological states of users in response to specific issues or events, researchers can probe the composition of stance communities to better understand why users hold stances, and whether they connect or interact with others who share similar views on social media. For example, it might be possible to identify communities of users expressing identical stances who also share demographic or psychometric similarities. Understanding such stance communities can aid in early detection of social fractures (e.g., by sensing the formation of social fault lines) and contribute to intervention efforts (e.g., through targeted interventions at the community level) to mitigate adverse societal outcomes.

### 5.4. Improving design of LLM-based models

Even prior to the advent of LLMs, other pre-trained language models, such as BERT based models (Vamvas & Sennrich, 2020), had demonstrated success in zero-shot cross-language and cross-task stance detection, although some performance gaps were reported. Other studies have highlighted that overfitting current models to popular stance detection datasets can introduce systematic biases (e.g., Schiller et al., 2021). This is also in line with findings by Alturayeif et al. (2023) who pointed out the potential for annotation bias in crowdsourced stance annotations, and called for more 'non-intrusive' data collection strategies.

LLMs offer a promising opportunity to tackle several of these challenges. Our discussion in Sec. 4 on the use of LLMs for stance detection has highlighted recent successes with techniques such as agent-based/role-infused architectures and/or CoT reasoning in enhancing stance detection tasks. For example, Cruickshank & Ng (2024) compared the effectiveness of 10 large language models and 7 different prompting schemes on stance detection tasks based on CoT and collaborative role-infused prompt designs (e.g., Lan et al., 2024). Recent studies have extended the design of collaborative expert agents to a dynamic framework (e.g., Wang et al., 2024) where a pool of experts is first constructed from the labelled sample. Then experts are dynamically selected to reason about a particular sentence, and the stance is subsequently inferred. Recent models have also explored ways to better incorporate contextual information about the target. For instance, Li et al. (2023) proposed the use of contextual knowledge via episodic knowledge from open repositories (e.g., Wikipedia) and discourse knowledge (via content expansion) to improve performance on zero-shot stance detection tasks.

The problem of mitigating prediction bias is also an emerging area of work. For example, Li et al. (2024) identified two types of biases in LLM-based stance detection tasks – sentiment-stance correlations and target bias. The former refers to biased stance predictions driven by the sentiment expressed in the text, while the latter refers to inherent biases the LLM may have towards specific targets. The authors proposed a calibration network to do post-hoc adjustment of LLM-generated stances to correct for these biases. Lastly, in addition to generating stance labels, LLM-based models can also generate justification texts to enhance the interpretability of the stance predictions (e.g., memory network model; Mohtarami et al., 2018).

## 6. CONCLUSION

Social media platforms offer users the perfect venue to express opinions on a variety of topics, and in response to emerging events of interest. While the task of detecting stances from user generated content has gained considerable interest in the past decade, several critical gaps remain. This paper highlights the need to adopt a more theoretically-informed approach to modelling stance, particularly at a user level, by leveraging related attributes that offer predictive value in such tasks. While previous research has disproportionately focused on message-level stance detection tasks, the advent of language models, and particularly LLMs, offer a significant opportunity for researchers to refocus on *user-level stance detection* problems. Specifically, we argue that this can be achieved by modelling and inferring deeper user characteristics (e.g., psychological attributes) using language models, which can be subsequently incorporated in detecting user stances. With this, we foresee an increase in the number of studies that infer user-level stances in high value online contexts such as perceptions towards emerging events (e.g., elections, climate change, new policies). Consequently, this will also lead to improved inclusivity of stance research through the development of large, user-level, and stance-labelled datasets in multiple languages.

## ACKNOWLEDGMENTS

This research is supported by the SMU-A*STAR Joint Lab in Social and Human-Centered Computing (SMU grant no.: SAJL-2022-CSS02, SAJL-2022-CSS003). This research is supported by A*STAR (C232918004, C232918005).

**APPENDIX**

**Table A1: An interdisciplinary synthesis of stance definitions and their implications for computational modelling**

| Source/authors | Reference discipline | Illustrative definitions | Diagnostic themes | Relevant data modalities |
|---|---|---|---|---|
| **General definitions** | | | | |
| Biber and Finegan (1988) | Linguistics | "By stance we mean the lexical and grammatical **expression** of **attitudes**, **feelings**, **judgments**, or **commitment** concerning the **propositional content** of a message." | (i) Public expression, (ii) Cognitive and affective factors, (iii) Target specific | (i) Semantic (text, speech), (ii) Psychological (e.g., attitudes, moral foundations, beliefs) |
| Biber et al. (1999) | Linguistics | "..**personal feelings**, **attitudes**, **judgments**, or **assessments** that a speaker or writer has about the information in a **proposition**" | (i) Cognitive and affective factors (ii) Target specific | (i) Semantic (text, speech), (ii) Psychological (e.g., attitudes, moral foundations, beliefs) |
| Biber (2004) | Linguistics | "**..**stance is the expression of one's **personal viewpoint** concerning p**roposed information**" | (i) Public expression, (ii) Cognitive factors (iii) Target specific | (i) Semantic (text, speech), (ii) Psychological (e.g., attitudes, moral foundations, beliefs) |
| Teller (2004) | Philosophy | "A **philosophical position** can consist in a stance (**attitude**, **commitment**, approach, a cluster of such – possibly including some **propositional attitudes** such as **beliefs** as well). Such a stance can of course be expressed, and may involve or presuppose some beliefs as well, but cannot be simply equated with having beliefs or **making assertions** about what there is" | (i) Cognitive factors - distinction between stance and beliefs (ii) Target specific | (i) Psychological (e.g., attitudes, moral foundations, beliefs) |
| Kockelman (2004) | Linguistics | "Stance may be understood as the semiotic means by which we indicate our **orientation** to **states of affairs**, usually framed in terms of **evaluation** (e.g., **moral obligation** and **epistemic possibility**) or **intentionality** (e.g., **desire** and memory, **fear and doubt**)…a way of categorizing and judging experience particular | (i) Public expression, (ii) Cognitive factors (iii) Target specific (iv) Individual & collective stances | (i) Semantic (text, speech), (ii) Psychological (e.g., attitudes, moral foundations, beliefs) |

| | | to a **group or individual** that turns on **some notion of the good or true**. (Ash 2001; Helmers 1998)." | | |
|---|---|---|---|---|
| Kockelman (2004) | Linguistics, Anthropology | "As may be seen from these uses, stance is not so much a new topic in linguistics as it is a new name for what is often called the speaker's **attitude**, **view**, or **evaluation**…. In some sense, then, as the terms themselves indicate, the turn from **attitude** to stance is in keeping with other trends in linguistics and anthropology: from an emphasis on the **private, subjective, and psychological (attitude)** to an emphasis on the **public, intersubjective, and embodied (stance)**" | (i) Public expression, (ii) Cognitive factors (iii) Target specific (iv) Individual & collective stances | (i) Semantic (text, speech), (ii) Psychological (e.g., attitudes, moral foundations, beliefs) |
| Precht (2003) | Linguistics | "..the expression of **attitude, emotion, certainty and doubt**" | (i) Public expression, (ii) Cognitive and affective factors | (i) Semantic (text, speech), (ii) Psychological (e.g., attitudes, moral foundations, beliefs) |
| Kiesling (2009) | Sociolinguistics | "..stances are taken not by using a single variant, but with a range of **social practices** in the manner that the Stanford group outlines for **personal styles**. …The view that stances underlay these personal styles and **personae** is completely compatible with this view of style: we simply understand the personal styles to be repertoires of stances" | (i) Public expression via styles | (i) Communicative styles/behaviour (ii) Psychological (e.g., traits) |
| Kiesling (2009) | Sociolinguistics | "…we find that what tends to **differentiate census-like groups**— in the discourses of the society that define them, real or imagined—are the stances they **habitually** take.. Thus, **identity** and **personal style** are both ways of stereotyping habitual patterns of stance taking, or repertoires of stance" | (i) Public expression via styles (ii) Role of group identification | (i) Communicative styles/behaviour (ii) Social (group identity) (iii) Psychological (e.g., traits) |
| Boucher (2014) | Philosophy | "**…they are pragmatically justified perspectives** or ways of seeing the world…stances should be understood as pragmatically justified metaphysical perspectives, or **ways of seeing**. They are particular **orientations** on the world, or ways of seeing facts. Such ways of seeing are typically justified both in terms of their **epistemic** fruits, and in terms of their coherence with **one's values**" | (i) Cognitive factors (e.g., orientations, worldviews) (iii) Target specific | (ii) Psychological (e.g., primal beliefs, moral foundations, preferences) |
| Van Fraassen (1995) *(similar to Teller 2004)* | Philosophy | "…a **philosophical position** can consist in something other than a belief in what the world is like. The alternative is a stance (**attitude, commitment, approach**) which can be expressed, and which may involve or presuppose some **beliefs** as well … What empiricists have shared over the centuries… has not most obviously been a set of beliefs… [empiricism is] an **attitude**, or rather a **cluster of attitudes**, a philosophical stance." | (i) Cognitive factors - distinction between stance and beliefs (ii) Target specific | (i) Psychological (e.g., attitudes, moral foundations, beliefs) |
| Boucher (2014) | Philosophy | "Stances may cohere with either one's **epistemic or one's nonepistemic values**. Nonepistemic values may include **social,** | (i) Cognitive factors (ii) Socio-political factors | (i) Psychological (e.g., epistemic and non-epistemic values) (ii) Social (e.g., group identity) |

| | | political, ethical or aesthetic values. Epistemic values have to do rather with the search for truth" | | (iii) Cultural (e.g., aesthetic styles) |
|---|---|---|---|---|
| Chakravartty (2004) | Philosophy | "'A stance is a strategy, or a **combination of strategies**, for generating **factual beliefs**… It is ... a sort of epistemic **'policy'** concerning which methodologies should be adopted in the generation of factual beliefs… Stances are not themselves factual— they are possible means to realms of possible facts'" | (i) Public expression (ii) Cognitive factors | (i) Psychological (e.g., beliefs) |
| **Interpersonal stance** | | | | |
| Klaus Scherer (2005) | Social Psychology | "..it is characteristic of **an affective style** that **spontaneously** develops or is **strategically** employed in the interaction with a person or a group of persons, coloring the **interpersonal exchange** in that situation (e.g. being **polite, distant, cold, warm, supportive, contemptuous**). Interpersonal stances are often triggered by events, such as encountering a certain person, but they are less shaped by spontaneous appraisal than by **affect dispositions, interpersonal attitudes, and, most importantly, strategic intention**." | (i) Public expression (ii) Target specific (iii) Relevant to social interactions (iv) Cognitive and affective factors | (i) Communicative styles/behaviour (ii) Psychological (e.g., target-specific attitudes) (iii) Social networks (e.g., relationship type and strength) |
| Kiesling (2009) | Sociolinguistics | "…a person's **expression of their relationship** to their interlocutors (their interpersonal stance—e.g., friendly or dominating)." | (i) Public expression (ii) Target specific (iii) Relevant to social interactions | (i) Communicative styles/behaviour (ii) Social networks (e.g., relationship type and strength) |
| | | | | |
| **Social stance** | | | | |
| Du Bois (2007) | Sociolinguistics | "a public act by a social actor, achieved **dialogically** through **overt communicative means** (language, gesture, and other symbolic forms), through which **social actors** simultaneously evaluate objects, position subjects (themselves and others), and align with other subjects, with respect to any salient dimension of the sociocultural field." | (i) Public expression via verbal and nonverbal modes (ii) Target specific (iii) Relevance of socio-cultural context | (i) Semantic (text, speech, visual, physiological), (ii) Social networks (e.g., relationship types) (iii) Cultural (e.g., social norms) |
| Du Bois (2007) | Sociolinguistics | "..an **articulated** form of **social** action" | (i) Public expression (ii) Relevance of socio-cultural context | (i) Semantic (text, speech, visual, physiological) |
| **Epistemic stance** | | | | |
| Kiesling (2009) | Sociolinguistics | "..stance as a person's expression of their relationship to their talk (their epistemic stance—e.g., how **certain they are about their assertions**)" | (i) Public expression (ii) Certainty of assertions | (i) Semantic (text, speech, visual, physiological) |
| Teller (2004) | Philosophy | "…the idea of a stance should be taken to be an open-ended notion, and insofar as it is positively characterized, **characterized functionally**. We need guidelines for ways to form and **evaluate** | (i) Public expression (ii) Cognitive factors (iii) Target specific | (i) Psychological (e.g., attitudes, beliefs) (iii) Cultural (e.g., social norms) |

| | | **beliefs**. An epistemic stance is such a guideline. To characterize an epistemic stance as an epistemic guide is to characterize it functionally, in terms of what it does…. to operate as an epistemic guide, but not as a thesis, a stance will be a matter of **personal decision and commitment**, if only by default by being presumed by the **traditions of one's community**" | (iv) Relevance of socio-cultural context | |
|---|---|---|---|---|
| Biber et al. (1999) | Linguistics | "Epistemic markers express the **speaker's judgment** about the **certainty, reliability, and limitations** of the proposition; they can also comment on the source of the information." | (i) Public expression (ii) Identifiable through linguistic markers (iii) Target specific | (i) Semantic (text, speech, visual, physiological) |
| **Affective stance** | | | | |
| Chindamo et al. (2012) | Social computing / Social signal processing | "..while affective stance is related to the **emotional feelings** about the object of **discourse**" | (i) Private experience and/or public expression (ii) Affective factors (iii) Target specific | (i) Sentiment and emotions (text, speech, visual and physiological) |
| Ochs (1996) | Sociolinguistics | "a **mood**, **attitude**, **feeling** and **disposition**, as well as degrees of **emotional intensity** vis-a.-vis some focus of concern." | (i) Private experience and/or public expression (ii) Affective factors (iii) Target specific | (i) Sentiment and emotions (text, speech, visual and physiological) |
| Goodwin et al. (2012) | Sociolinguistics, Communication | "..display of **emotion** is a situated practice entailed in a speaker's performance of affective stance through **intonation, gesture, and body posture**" | (i) Public expression via non-verbal cues (ii) Affective factors (iii) Target specific | (i) Sentiment and emotions (text, speech, visual and physiological) |
| Voutilainen et al. (2014) | Linguistics (pragmatics), Communication | "..the term stance refers to the teller's **affective treatment** of the events he or she is talking about, or in a broad sense, to the **emotional valence** of the events as expressed by the teller" | (i) Public expression via non-verbal cues (ii) Affective factors (iii) Target specific | (i) Sentiment and emotions (text, speech, visual and physiological) |
| **Attributed stances - Intentional stance** | | | | |
| Robbins & Jack (2006) | Philosophy | "..denote the capacity to ascribe **intentional mental states** and to predict and explain behavior on the basis of those ascriptions" | (i) Attributing intentions to target-specific behaviour | (i) Communicative styles/behaviour (ii) Psychological (e.g., target-specific attitudes, beliefs) (iii) Cultural (e.g., norms, construals) (iv) Social network (e.g., relationship between observer and target) |
| Robbins & Jack (2006) | Philosophy | "To adopt the intentional stance toward X is to understand X as an intentional system, that is, to regard X as a locus of intentionality. This entails **ascribing intentional states (beliefs,** | (i) Attributing intentions and associated cognitive factors to | (i) Communicative styles/behaviour (ii) Psychological (e.g., target-specific attitudes, beliefs) |

| | | desires, intentions, etc.) to X** and using those ascriptions to make sense of X's behavior." | target-specific behaviour | (iii) Cultural (e.g., norms, construals)<br>(iv) Social network (e.g., relationship between observer and target) |
|---|---|---|---|---|
| Dennett (1981) | Philosophy | "One adopts the intentional stance toward any system one **assumes to be (roughly) rational** -- to adopt a **truly moral stance** toward the system (thus viewing it as a person) might often turn out to be psychologically irresistible given the first choice, but it is logically distinct. We might, then, distinguish a fourth stance, above the intentional stance, called the personal stance. The personal stance presupposes the intentional stance -- and seems, to cursory view at least, to be just the **annexation of moral commitment to the intentional**" | (i) Attributing intentions and associated cognitive factors to target-specific behaviour | (i) Communicative styles/behaviour<br>(ii) Psychological (e.g., target-specific attitudes, beliefs)<br>(iii) Cultural (e.g., norms, construals)<br>(iv) Social network (e.g., relationship between observer and target) |
| **Phenomenal stance** | | | | |
| Robbins and Jack (2006) | Philosophy | "Analogously, to adopt the phenomenal stance toward X is to understand X as a 'phenomenal system,' that is, to regard X as a locus of **phenomenal experience**…Part of what it is to regard something as a locus of experience is ascribing **phenomenal states (emotions, moods, pains, visual sensations**, etc.) to it. But this involves more than mere rote ascription of phenomenal states; it requires a felt appreciation of their **qualitative character**." | (i) Attributing intentions and associated cognitive and affective factors to target-specific behaviour | (i) Communicative styles/behaviour<br>(ii) Psychological (e.g., target-specific attitudes, beliefs)<br>(iii) Cultural (e.g., norms, construals)<br>(iv) Social network (e.g., relationship between observer and target) |

**Table A2: Summary of LLM-based stance detection studies**

| Source | Authors | Task/problem | Definition/stance classes | Domain/context | Unit of analyses | Key approach | Methods / evaluation |
|---|---|---|---|---|---|---|---|
| NAACL | Zhang et al. (2024a) | Zero-shot stance detection and cross-target stance detection | Support, against, neutral | Social media (datasets: P-Stance, VAST) | Post | LLM-enhanced modelling pipelines | Used partially filled zero-shot prompts to extract LLM-driven knowledge for stance generation with a BART-based decoder model. |
| Prepint (arXiv) | Zhang et al. (2023b) | Zero-shot stance detection | Support, against, neutral | Social media (datasets: SemEval-2016 Task 6, VAST) | Post | Novel prompting schemes | Proposed a logically consistent CoT for performing zero-shot stance detection |
| ArgMining-WS 2023 | Sharma et al. 2023 | Argumentative stance prediction | Support, against, | Social media (dataset: ImgArg) | Image-text pair corresponding to a tweet | LLM-enhanced modelling pipelines | Combined an ensemble of fine-tuned LLMs through a weighted sum of their individual predictions. Used InstructBLIP-based image summarization to augment original tweets for RoBERTa classification. |
| ICASSP | Ding et al. (2024a) | Cross-target stance detection | Support, against, neutral | Social media (datasets: SemEval-2016 Task 6, VAST) | Post | LLM-enhanced modelling pipelines | Applied a two-stage chain-of-thought method for eliciting target analysis perspectives and generating natural |

| Source | Authors | Task/problem | Definition/stance classes | Domain/context | Unit of analyses | Key approach | Methods / evaluation |
|---|---|---|---|---|---|---|---|
| | | | | | | | language explanations (NLEs). Developed a multi-perspective prompt tuning framework to integrate NLEs into the stance predictor |
| EMNLP | Kang et al. (2023) | Opinion prediction | Degrees of agreement (0 to 10) | Media and Social Trust, Personal and Social Well-Being, Politics (dataset: ESS) | User | Fine-tuned LLMs | Target value distributions were injected into LLMs via fine-tuning (argument generation and question answering) to predict opinions and behaviours of people with similar value distributions |
| EMNLP | Gatto et al. (2023) | Stance detection | Support, against, neutral | Social media (datasets: Tweet-Stance, Presidential-Stance) | Post | LLM-enhanced modelling pipelines | Employed a 1-shot Chain-of-Thought (CoT) prompting for stance detection on tweets. Integrated CoT reasoning from the LLM with RoBERTa embeddings, and incorporated it into the stance detection pipeline. |
| AAAI | Lan et al. (2024) | Stance detection | Support, against, neutral | Social media (datasets: SemEval-2016 | Post | LLMs as multi-agent systems | Utilized three LLMs as distinct experts—a linguistic expert, a domain specialist, |

| Source | Authors | Task/problem | Definition/stance classes | Domain/context | Unit of analyses | Key approach | Methods / evaluation |
|---|---|---|---|---|---|---|---|
| | | | | Task 6, VAST, P-Stance) | | | and a social media veteran—to provide a multifaceted analysis of texts. Leveraged the LLMs to detect logical connections between text features and three stances (Pro, Con, and Neutral). Derived the final stance judgment through an LLM based on the outputs from the previous steps |
| Preprint (arXiv) | Cruickshank & Ng (2024) | Stance detection | For, against, neutral, unrelated | Social media (datasets: covid-lies, election2016, phemerumors, SemEval2016, srq, wtwt) | Post | LLM evaluations using crowd-sourced annotations and other models | Evaluated stance detection performance of 10 open-source LLMs using 7 prompting schemes, with Few-Shot Prompting (FSP) and Zero-Shot Chain-of-Thought (CoT) demonstrating superior results. |
| Preprint (arXiv) | Li et al. (2024) | Stance detection | Support, against, neutral | Social media (datasets: covid-lies, | Post | LLM-enhanced | Prompted LLMs to rephrase original sentences with varied |

| Source | Authors | Task/problem | Definition/stance classes | Domain/context | Unit of analyses | Key approach | Methods / evaluation |
|---|---|---|---|---|---|---|---|
| | | | | election2016, phemerumors, SemEval2016, srq, wtwt) | | modelling pipelines | words and sentiments or reversed the stance of the original samples to generate counterfactual examples, introducing disturbances to non-causal features.<br><br>Encoded LLM outputs with BERT and passed the embeddings to a gated calibration network |
| Preprint (arXiv) | Li & Conrad (2024) | Stance detection | Favor, oppose, neutral, irrelevant | Social media (dataset: curated dataset) | Post | LLM evaluations using crowd-sourced annotations and other models | Zero-shot and few-shot prompting for stance detection; compared/evaluated results with human annotators' judgments. |
| Preprint (arXiv) | Yang et al. (2024) | Stance detection | Support, Deny, Question, Comment | Social media (datasets: RumorEval-S, SemEval-8) | Post | Fine-tuned LLMs | Reinforcement tuning framework applied to two LLMs for joint stance detection and rumor verification tasks. |

| Source | Authors | Task/problem | Definition/stance classes | Domain/context | Unit of analyses | Key approach | Methods / evaluation |
|---|---|---|---|---|---|---|---|
| Preprint (arXiv) | Gatto et al. (2024) | Stance detection | In favor, against, none | Social media (dataset: Long COVID-Stance) | Post | LLM evaluations using crowd-sourced annotations and other models | Evaluated stance detection performance of three LLMs (Llama2, GPT-3.5, and GPT-4) using three prompting strategies: zero-shot, few-shot, and chain-of-thought (CoT). |
| Preprint (arXiv) | Aiyappa et al. (2024) | Zero-shot stance detection | Support, against, neutral | Social media (datasets: SemEval-2016 Task 6A6B, P-Stance) | Post | LLM evaluations using crowd-sourced annotations and other models | Evaluated zero-shot stance detection performance of FlanT5-XXL using various prompts and decoding strategies. |
| Plos one | Mets et al. (2024) | Stance detection | Support, against, neutral | Online political discussion (dataset: immigration-related and curated dataset) | Post | LLM evaluations using crowd-sourced annotations and other models | Evaluated zero-shot prompted GPT-3.5 against other fine-tuned models on the curated immigration dataset. |
| Preprint (arXiv) | Wagner et al. 2024 | Stance detection | Support, against | Online political discussion (dataset: German comments in X- | Post | LLM-enhanced modelling pipelines | Prompted Mistral-7B to generate synthetic comments based on given political |

| Source | Authors | Task/problem | Definition/stance classes | Domain/context | Unit of analyses | Key approach | Methods / evaluation |
|---|---|---|---|---|---|---|---|
| | | | | Stance dataset) | | | questions and stances.<br><br>Proposed a Synthetic Data-driven Query by Committee (SQBC) method to identify the most informative samples in the unlabelled data pool by selecting samples with the most indecisive scores based on k-nearest neighbors among the synthetic data. |
| LREC-COLING | Niu et al. (2024) | Conversational stance detection | Favor, against, none | Social media (dataset: MT-CSD) | Post | LLM evaluations using crowd-sourced annotations and other models | Compared proposed method, global-local attention network (GLAN), with one-shot prompted ChatGPT and LLaMA (as well as other DNNs and PLMs). |
| LREC-COLING | Wang et al. (2024) | Stance detection | Support, against, neutral | Social media (datasets: SemEval-2016, MSTD, P-Stance) | Post | LLMs as multi-agent systems | Few-shot prompted the LLM to select from *manually created* experts for stance detection. |

| Source | Authors | Task/problem | Definition/stance classes | Domain/context | Unit of analyses | Key approach | Methods / evaluation |
|---|---|---|---|---|---|---|---|
| | | | | | | | Subsequently, used *LLM-generated experts* via in-context learning to reason and discuss the stance, discarding experts with low prediction accuracy and frequency, and retrieving the top-h related experts based on textual similarity scores from a sentence-expert repository. |
| Mathematics | Dong et al. (2024) | Stance detection | Support, against, neutral | Social media (datasets: SemEval-2016, ISD) | Post | LLM-enhanced modelling pipelines | Employed a zero-shot prompt to extract hashtag knowledge from the LLM, used the BERT model to generate embedding vectors for the additional hashtag knowledge, and fed these embeddings into a fusion network for stance detection. |
| LREC-COLING | Reuver et al. (2024) | Few-shot cross-topic stance detection | two stance task definitions (Pro/Con, Same Side Stance) | Social media (datasets: arc, semeval2016, argmin, iac, | Post | Fine-tuned LLMs | Evaluated the robustness of operationalization choices (task |

| Source | Authors | Task/problem | Definition/stance classes | Domain/context | Unit of analyses | Key approach | Methods / evaluation |
|---|---|---|---|---|---|---|---|
| | | | | scd, ibmcs, perspectrum) | | | definition, architecture, and pre-fine tuning on NLI) for few-shot cross-topic stance detection |
| Computational Linguistics | Ziems et al. (2024) | Stance detection | Support, against, none | Social media (dataset: SemEval-2016) | Post | LLM evaluations using crowd-sourced annotations and other models | Evaluated zero-shot prompted FLAN-T5, GPT-3, GPT-4, and other models on stance detection task. |
| Preprint (arXiv) | Gül et al. 2024 | Stance detection | Support, against, none | Social media (datasets: SemEval-2016, P-Stance, Twitter Stance Election 2020) | Post | Fine-tuned LLMs | Employed fine-tuned ChatGPT, LLaMa-2, and Mistral-7B for stance detection, surpassing existing benchmarks and highlighting their efficiency in zero-shot and few-shot learning scenarios. |
| IEEE Transactions on Computational Social Systems | Zhang et al. (2024d) | Stance detection | Support, against, neutral | Social media (datasets: SemEval-2016, VAST, P-Stance) | Post | LLM-enhanced modelling pipelines | Developed a knowledge-augmented interpretable network for zero-shot stance detection, which utilized LLM analysis to bridge connections |

| Source | Authors | Task/problem | Definition/stance classes | Domain/context | Unit of analyses | Key approach | Methods / evaluation |
|---|---|---|---|---|---|---|---|
| | | | | | | | between seen and unseen targets and integrated this knowledge through a bidirectional knowledge-guided neural system for stance detection. |